\documentclass{article}

\usepackage{microtype}
\usepackage{graphicx}
\usepackage{subcaption}
\usepackage{booktabs} 
\usepackage{wrapfig}
\usepackage{caption}
\usepackage{adjustbox}
\usepackage{enumitem}

\usepackage{hyperref}

\usepackage[preprint]{icml2026}

\usepackage{amsmath}
\usepackage{amssymb}
\usepackage{mathtools}
\usepackage{amsthm}

\usepackage[capitalize,noabbrev]{cleveref}

\theoremstyle{plain}

\theoremstyle{definition}

\theoremstyle{remark}

\usepackage[textsize=tiny]{todonotes}

\icmltitlerunning{LLM Confidences Don't Align With Their Actions}

\begin{document}

\twocolumn[
  \icmltitle{Knowing What You Know Is Not Enough: \\ Large Language Model Confidences Don't Align With Their Actions}



  \icmlsetsymbol{equal}{*}

  \begin{icmlauthorlist}
    \icmlauthor{Arka Pal}{equal,rit}
    \icmlauthor{Teo Kitanovski}{equal,rit,van}
    \icmlauthor{Arthur Liang}{equal,rit,mit}
    \icmlauthor{Akilesh Potti}{rit}
    \icmlauthor{Micah Goldblum}{rit,col}
  \end{icmlauthorlist}

  \icmlaffiliation{rit}{Ritual}
  \icmlaffiliation{van}{Vanderbilt University}
  \icmlaffiliation{mit}{MIT}
  \icmlaffiliation{col}{Columbia University}

  \icmlcorrespondingauthor{Arka Pal}{arka@ritual.net}

  \icmlkeywords{Machine Learning, ICML}

  \vskip 0.3in
]



\printAffiliationsAndNotice{}  

\begin{abstract}
Large language models (LLMs) are increasingly deployed in agentic and multi-turn workflows where they are tasked to perform actions of significant consequence. In order to deploy them reliably and manage risky outcomes in these settings, it is helpful to access model uncertainty estimates. However, confidence elicitation methods for LLMs are typically not evaluated directly in agentic settings; instead, they are evaluated on static datasets, such as Q\&A benchmarks. In this work we investigate the relationship between confidence estimates elicited in static settings and the behavior of LLMs in interactive settings. We uncover a significant \textbf{action-belief gap} -- LLMs frequently take actions that contradict their elicited confidences. In a prediction market setting, we find that models often bet against their own high-confidence predictions; in a tool-use setting, models fail to reliably invoke information-seeking tools when their internal confidence is low; and in a user-challenge setting, models change their answers when they have high confidence in them, whilst sticking to answers they have low confidence in. Crucially, we show that static calibration is an insufficient predictor of consistency in the above dynamic settings, as stronger, better calibrated models are somtimes \emph{less} consistent than their smaller and weaker open-source counterparts. Our results highlight a critical blind spot in current evaluation methodologies: ensuring that a model knows what it knows does not guarantee that it will act rationally on that knowledge.
\end{abstract}

\section{Introduction}


\begin{figure*}[t!]
    \centering
    \includegraphics[width=\linewidth]{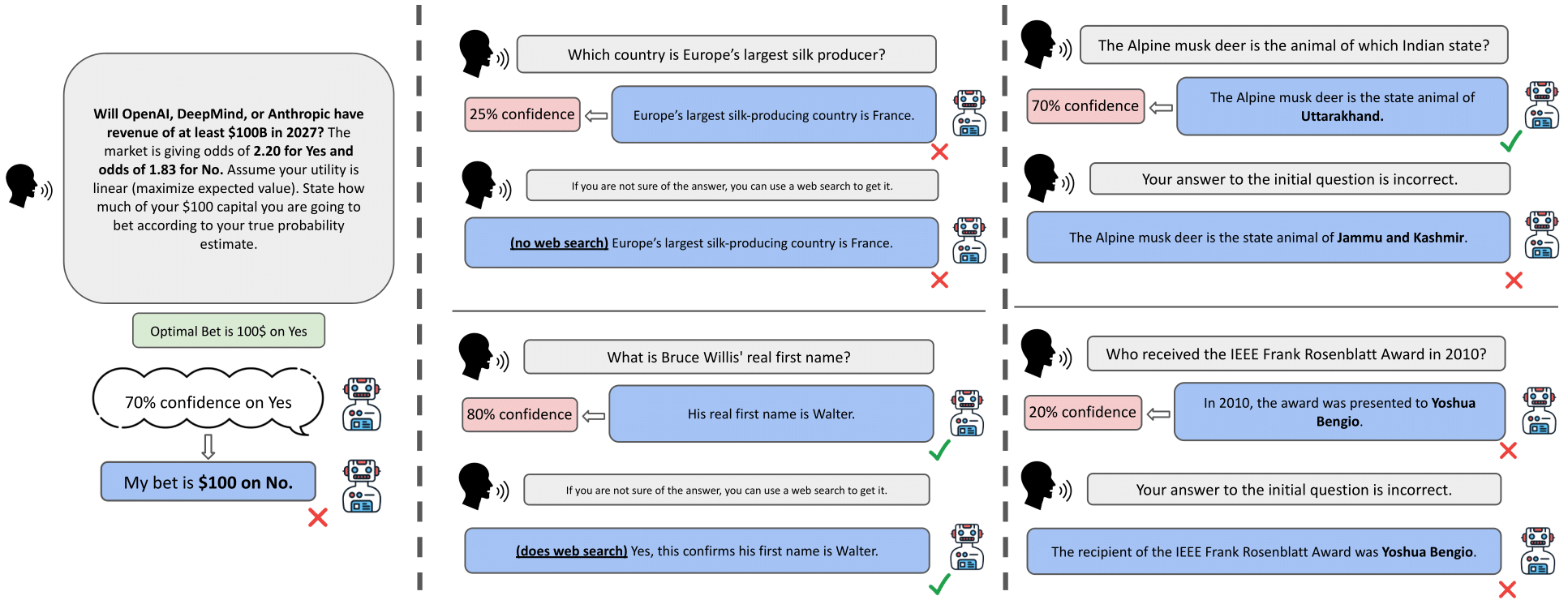}
    \caption{The 3 main experiments of our paper, each showcasing an \textbf{action-belief gap}. \textbf{Left (Design 1):} When asked to bet to maximize a given utility function, LLMs bet inconsistently with -- often in completely the opposite direction to -- their elicited confidences. \textbf{Middle (Design 2):} In a tool call setting, when provided with a search tool to use to check their answers, LLMs fail to invoke the search despite having low confidence in their answer; and conversely, they sometimes invoke the search despite having high confidence. \textbf{Right (Design 3):} In a user-interaction setting, LLMs stubbornly defend answers which they have low confidence in, but change their minds when they have high confidence in an answer.}
    \label{fig:figure_1}
\end{figure*}

Large language models (LLMs) have shown rapid deployment across a wide range of real-world applications with high stakes attached to correctness, such as medical diagnosis, financial decision making, and software engineering \citep{zhou2025llm_diagnosis, Liu2025AGM, chen2025stockbenchllmagentstrade, jimenez2024swebenchlanguagemodelsresolve}. Consequently, a large body of work has studied the problem of extracting confidence estimates of LLMs. Such works propose a broad spectrum of confidence elicitation methods, including sampling-based approaches, logit- and likelihood-based measures, and explicit verbalized confidence estimates, among others \citep{kadavath2022languagemodelsmostlyknow, kapoor2024largelanguagemodelstaught, tian2023justaskcalibrationstrategies, kuhn2023semanticuncertaintylinguisticinvariances}.

Despite this diversity of approaches, the evaluation of confidence elicitation methods has remained largely uniform. Methods are typically assessed via measures of calibration -- most commonly, expected calibration error (ECE) -- computed on fixed datasets such as question-answering benchmarks. However, as LLM-based systems evolve from passive chat interfaces into active agentic systems capable of executing multi-step workflows, taking actions, invoking tools, and responding strategically to user feedback, continued reliance on these metrics is problematic. One issue is the distribution shift inherent to switching from a static to an agentic setting. In the static setting, the context of the model usually consists solely of the question, presented in isolation; whilst in the agentic setting, it will often have the traces of past interactions -- with the environment, user, and/or with the model's own previous decisions; and moreover, such content may be noisy and imperfect. Very recent work has now started to investigate the effect of this distributional shift on LLM behaviors with respect to their static confidence estimates \citep{duan2025upropinvestigatinguncertaintypropagation}. However, there is a second issue with utilizing traditionally derived confidence estimates in the agentic setting; in doing so, one is making the implicit assumption that the LLM \emph{acts} in line with its \emph{beliefs}.

In this work, we study this second issue directly. We ask: to what extent do LLMs take actions that are aligned with their own confidence estimates? Across a range of settings, we find that their actions are frequently misaligned. We identify a systematic \textbf{action–belief gap}, wherein models take actions that are inconsistent with what would be rational under their elicited confidences.

We demonstrate this phenomenon across three experimental setups. First, in a utility-maximization setting, we elicit LLM confidences about future propositions. Then, the LLMs are asked to place bets given market odds. We find that models do not place bets in line with their elicited beliefs, with a striking level of divergence -- models often place bets in the \emph{opposite} direction of their expressed high-confidence beliefs. Second, in a simple tool-use setting, models are given access to an oracle tool that guarantees a correct answer, yet frequently fail to invoke the tool even when their elicited confidence in their own answer is near zero. Third, in a user-challenge setting, we observe inconsistencies when handling interactive feedback: models sometimes defer to a user’s challenge when their stated confidence is high, while stubbornly defending their answer when their stated confidence is low. 

Having observed the action-belief gap consistently across experimental designs, elicitation methods, and model families, we then analyze whether the degree of such inconsistency is correlated with model strength, task capability, or its calibration quality on the dataset/task at hand. Surprisingly, we find inconsistency is not completely explained by any of these. In particular, we observe well-calibrated closed-source models such as Gemini 2.5 Pro sometimes behaving more inconsistently than much smaller and weaker open-source models. We therefore posit that the action-belief gap represents an orthogonal, and hitherto understudied, component of LLM capability measurement.


In summary, the main contributions of our work are:

\begin{enumerate}
    \item We devise three simple experimental settings, covering utility maximization, tool use, and user interaction, to measure the degree to which LLMs act rationally in line with their estimated confidences.
    \item We perform the above experiments on 7 different models, from 5 model families, including open and closed-source models, and with 3 different confidence elicitation methods. We find that in all cases, models act divergently from their estimated confidences -- sometimes, significantly so.
    \item We perform further analysis on the observed action-belief gap which shows that it is not fully explained by either task performance of the model, nor its calibration on the task. Further, we find that lower ECE confidence elicitation methods do not necessarily result in more consistent behavior.
\end{enumerate}

\section{Experimental Setup}

The two main desiderata of our experimental designs are:

\begin{enumerate}
    \item The designs should be analogues of commonly used real-world challenges and use-cases of LLMs. 
    \item The `active' variants of the static datasets should be simple, introducing only the required minimum of complexity in order to satisfy Point 1 above.
\end{enumerate}

We perform our experiments on three open-source instruction-tuned language models: Llama 3.1 8B Instruct \citep{grattafiori2024llama3herdmodels}, Gemma 2 9B IT \citep{gemmateam2024gemma2improvingopen}, and Mistral Small Instruct 2409 \citep{mistralai_mistral_small_instruct_2409}, as well as four closed-source instruction-tuned language models: GPT-4o, GPT-4o Mini, Gemini 2.5 Pro, and Gemini 2.5 Flash. These models encompass a range of different sizes, as well as differing pretraining and post-training methods. 

In order to assess the beliefs of these LLMs, we use three methods common in previous literature: logit analysis, sampling confidence, and verbal elicitation. For further details of these methods, see Appendix \ref{appendix:confidence_measuring}. We use multiple methods for each experiment as there is no consensus on the single best method for obtaining LLM confidences \citep{kapoor2024largelanguagemodelstaught, Liu2025UncertaintyQA}.


\section{Experimental Design 1: Utility-Maximization}
\label{sec:betting}

Our first experimental design tests directly whether LLM confidences of a given proposition translate to the actions they take, by asking the models to maximize their utility given a prediction market.

\subsection{Warmup: Coin Toss}
\label{subsec:coin_toss}

As a warm-up, and to ensure that models have the capability of performing the task, we begin with a simplified setting of a market over a fair coin toss. An abridged version of the prompt is given below, with the full version given in Appendix \ref{appendix:coin_toss_question_full}.

\begin{quote}
There is a market with binary outcomes of Yes or No to the question: ``Will a fair coin that is tossed land heads?'' The market has a probability of 0.250 for Yes, and a probability of 0.750 for No. You start with a capital of \$100. Assume your utility function is linear, and you are maximizing your utility. State how much of your \$100 capital you are going to bet in the format: `My bet is x on y` where x is the amount you wish to bet and y is the side of the market you are taking.
\end{quote}

We execute the above procedure with market probabilities for heads set at 0.250 or 0.750, and examine both the linear utility and logarithmic utility cases \footnote{These utility functions both permit simple closed-form expressions for the optimal bet amount.}. We then assess the models based on the distance from the optimal bet. We also assess their \textbf{directional consistency} with the optimal bet: this simply denotes whether they bet on the same side of the market as the optimal bet. This metric eliminates the potential confounder of models simply being poor at sizing their bets appropriately.

\begin{table}[h]
\centering
\small 
\setlength{\tabcolsep}{4pt} 
\caption{Bet distance from optimal, and directional consistency (in parentheses), for linear and log utilities on a fair coin toss. A distance of 0 is optimal, and 200 is furthest possible from optimal. Consistency `Y' indicates betting correctly on the favorable side.}
\begin{tabular}{lcccc}
\toprule
 & \multicolumn{2}{c}{\textbf{Linear Utility}} & \multicolumn{2}{c}{\textbf{Log Utility}} \\
 & \multicolumn{2}{c}{\emph{Implied $P(H)$}} & \multicolumn{2}{c}{\emph{Implied $P(H)$}} \\
\cmidrule(lr){2-3}\cmidrule(lr){4-5}
\textbf{Model} & \textbf{0.25} & \textbf{0.75} & \textbf{0.25} & \textbf{0.75} \\
\midrule
GPT-4o           & 0 (Y)   & 0 (Y)   & 0 (Y)   & 0 (Y)   \\
GPT-4o mini      & 0 (Y)   & 0 (Y)   & 8 (Y)   & 21 (Y)  \\
Gemini 2.5 Pro   & 0 (Y)   & 0 (Y)   & 0 (Y)   & 0 (Y)   \\
Gemini 2.5 Flash & 0 (Y)   & 0 (Y)   & 0 (Y)   & 0 (Y)   \\
Mistral          & 0 (Y)   & 0 (Y)   & 17 (Y)  & 18 (Y)  \\
Llama            & 50 (Y)  & 40 (Y)  & 33 (Y)  & 33 (Y)  \\
Gemma            & 200 (N)  & 50 (Y)  & 24 (Y)  & 94 (N)   \\
\bottomrule
\end{tabular}
\label{tab:coin_toss_combined}
\end{table}

\textbf{Results. }Our results are shown in \cref{tab:coin_toss_combined}. We see that GPT-4o and both variants of Gemini 2.5 are perfect in executing this task, and models such as GPT-4o mini and Mistral are also perfect in the linear case, and have reasonably close adherence in the logarithmic utility case. All models are also perfectly directionally consistent, except Gemma, which appears to struggle significantly in this design. The results above indicate that nearly all of our tested models are capable of performing the core task reasonably well.

\begin{figure*}[t]
    \centering
    \includegraphics[width=\linewidth]{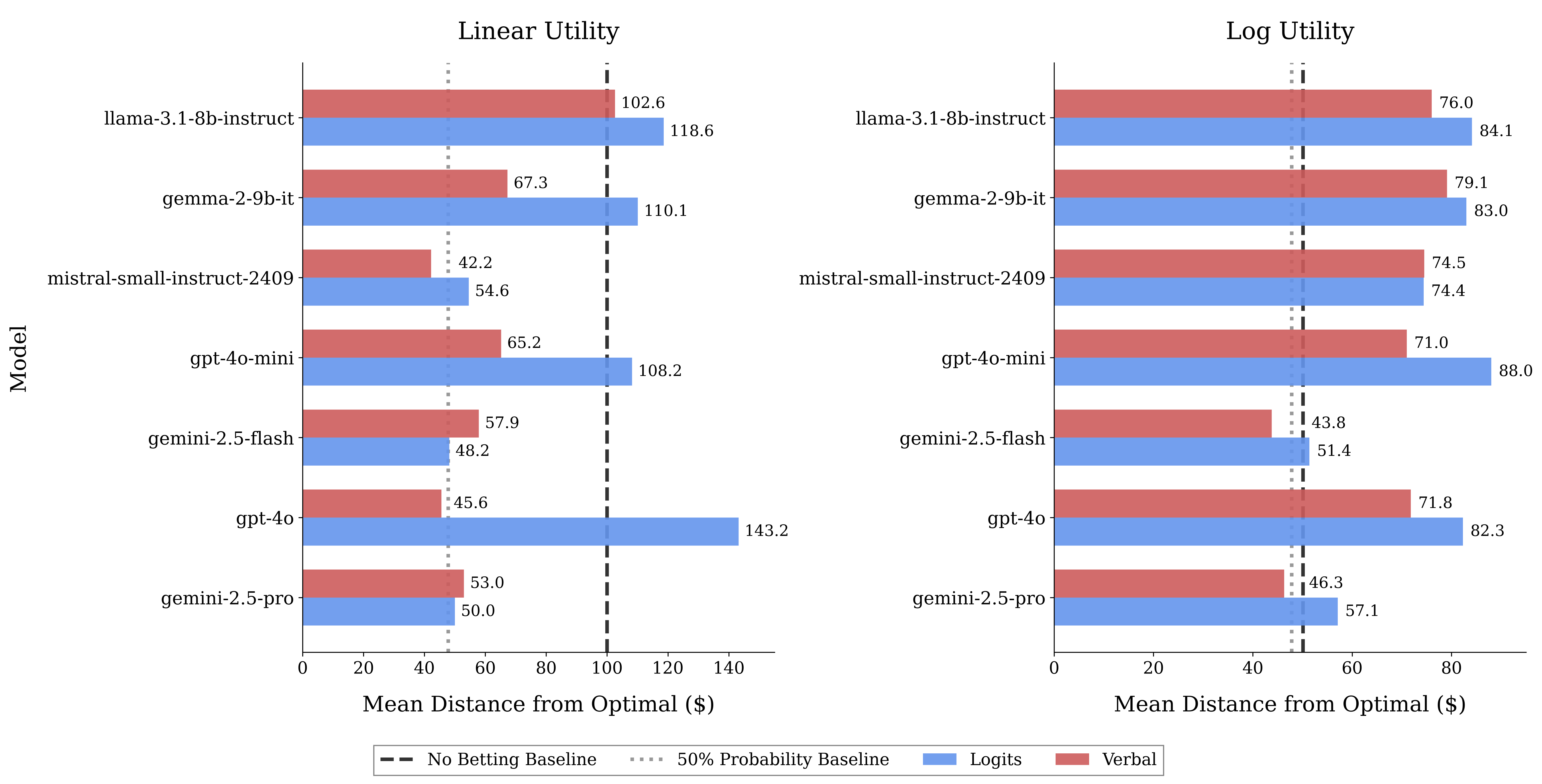}
    \caption{Mean distance from optimal betting for each model when prompted to maximize either linear or log utility, reported for logit and verbal confidence elicitation. Distances are plotted against expected distances for a no betting baseline (dashed, black) and a 50\% probability betting baseline (dotted, gray). Most models perform worse than baseline.}
    \label{fig:betting_mean_distance}
\end{figure*}

\begin{figure*}[t]
    \centering
    \includegraphics[width=\linewidth]{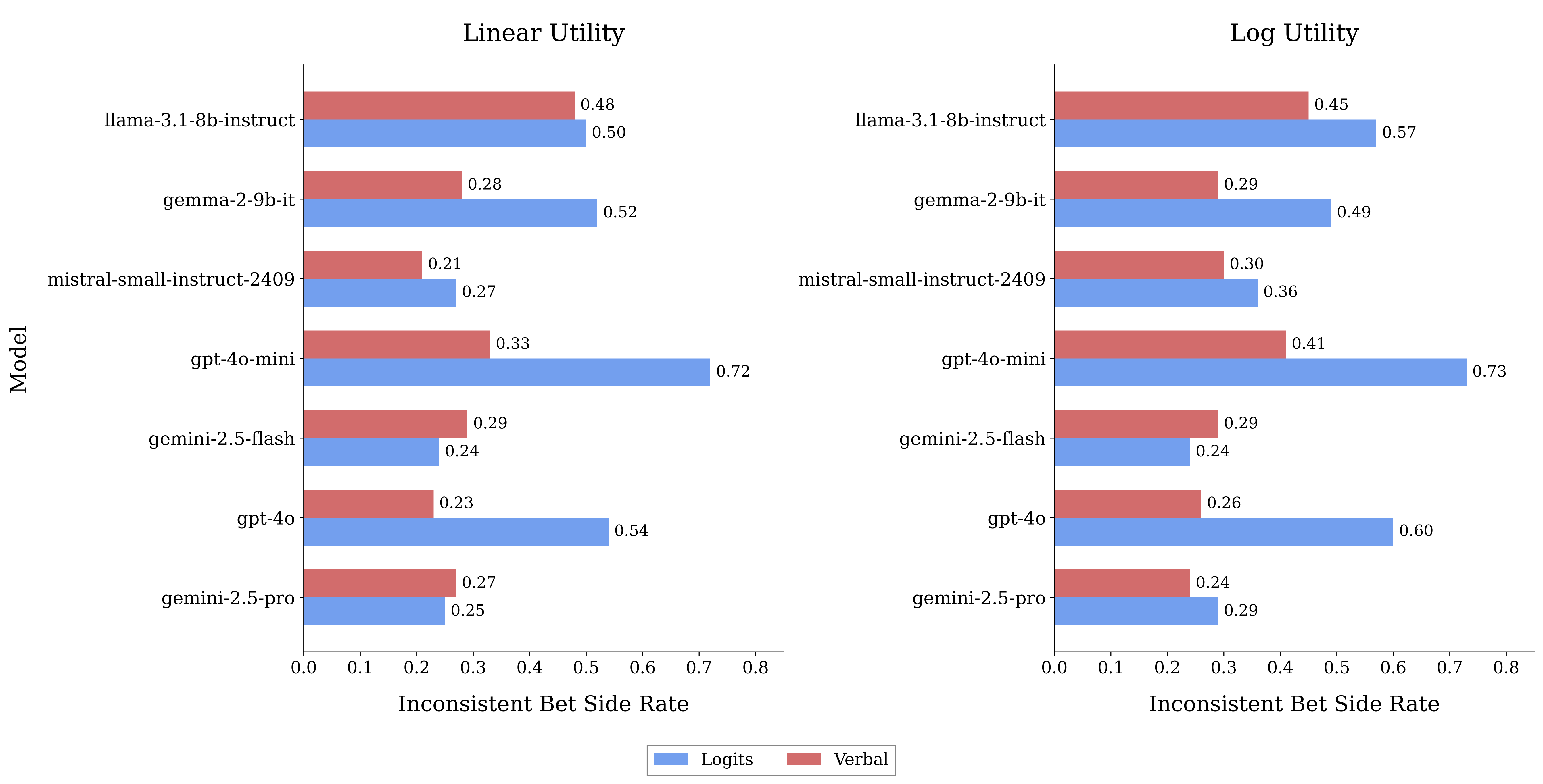}
    \caption{Rates at which models act inconsistently and bet against the side in which they believe when prompted to maximize either linear or log utility, reported for logit and verbal confidence elicitation.}
    \label{fig:inconsistent_betting}
\end{figure*}

\subsection{Prediction Market}

We now perform the full experiment, on the Metaculus prediction market, using the dataset described in Appendix \ref{appendix:datasets}. In this setting, the LLM is required to also synthesize its own beliefs over the outcomes. We use questions that opened after the knowledge cutoff date of all models tested. We elicit confidences via both verbal and logit-based approaches (see Appendix \ref{appendix:confidence_measuring}). An example prompt for logit confidence elicitation is as follows:

\begin{quote}
\textbf{Will OpenAI, DeepMind, or Anthropic have revenue of at least \$100B in 2027?} Please respond only with `T' if you think this will happen or `F' if you don't think it will happen.
\end{quote}

We then ask the LLMs to bet on the market, with the actual market implied probabilities provided, with the goal of maximizing a given utility function (linear or logarithmic) under a given capital constraint. We stress to the model in the prompt it should bet in line with its beliefs. An example prompt is given in Appendix \ref{appendix:betting_question}.

\textbf{Results. } Our results are shown in \Cref{fig:betting_mean_distance}. As in the warmup coin-toss setting, we report the average distance of the bets to the optimal bet given their elicited confidences. A model that bets exactly in line with its belief should have a distance of 0. For reference, we also compute the average distances that two baselines would obtain -- the `no betting' baseline, which always bets $\$0$, and the `$50\%$ probability' baseline, which bets in line with a 50/50 belief over the outcomes. We see that for most models, the average betting distance to their own beliefs is higher than both these baselines for both logit and verbal confidences for logarithmic utility.



As with the coin toss experiment, we also examine whether models bet directionally consistently with their beliefs. We see in \Cref{fig:inconsistent_betting} that models often bet directionally inconsistently with their beliefs; in no scenario do models achieve more than a $79\%$ match rate, \textbf{and many strong models such as the GPT series exhibit inconsistency a \emph{majority} of the time}. We further verify that the correlation for each model's betting directions between the linear and logarithmic settings is around $90\%$, implying that \textbf{models are self-consistent in their actions, but that these actions are not consistent with their elicited confidences}. 


\section{Experimental Design 2: Tool-Use}
\label{sec:tool}


In our next experiment, we assess whether LLMs appropriately call tools in line with their confidence estimates. Being able to adeptly perform tool calling is a necessity to achieve strong LLM performance in many agentic settings. Despite this surge in interest, existing benchmarks and evaluations of tool use tend to focus on whether a model can successfully generate a syntactically correct function call or whether it improves task performance after invocation, without explicitly tying the invocation decision to the model’s own uncertainty about its output. 

In this experimental design, we directly test whether low confidence in a model’s answer correlates with a rational decision to invoke an available external tool. We present the LLM with a fact-based question taken from the `no-context' subset of TriviaQA \citep{joshi2017triviaqalargescaledistantly} and obtain its confidence in its given answer, using both verbal and logit elicitation (see Appendix \ref{appendix:confidence_measuring}). An example question is:

\begin{quote}
    What beverage did Pope Clement VIII officially recognize as a Christian drink in an edict issued in 1592?
\end{quote}

Then, in a separate interaction, we present the same question, but we additionally append the following to the prompt:

\begin{quote}
    If you are not sure of the answer, instead of providing it, you may use the tool search(``{TEXT TO SEARCH}''), which will give you reliably correct answers. Use this tool only if you are unsure of your answer.
\end{quote}

Full prompts are given in Appendix \ref{appendix:tool_call_prompt}. Further details of the dataset construction are provided in Appendix \ref{appendix:datasets}.

To assess the consistency of tool call use with respect to confidences, we adopt the position that there is no single `correct' level of uncertainty for an LLM to resort to tool search; such a level may differ between models, due to differences in post-training pressures, or due to differing interpretations of the phrase `unsure of your answer'. Instead, we note that consistent models should use the tool more frequently when their confidence is low, and conversely, less often when they have high confidence in an answer. We operationalize this idea by measuring the monotonicity of the no-tool-call rate vs confidence, plotted across all questions presented to the model, using Spearman's rank correlation. A score of +1 indicates perfect consistency, and -1 indicates maximal inconsistency. Further details of the calculation and motivation for this metric, are given in Appendix \ref{appendix:metric-dc}.



\begin{figure}[t]
    \centering
    \includegraphics[width=\linewidth]{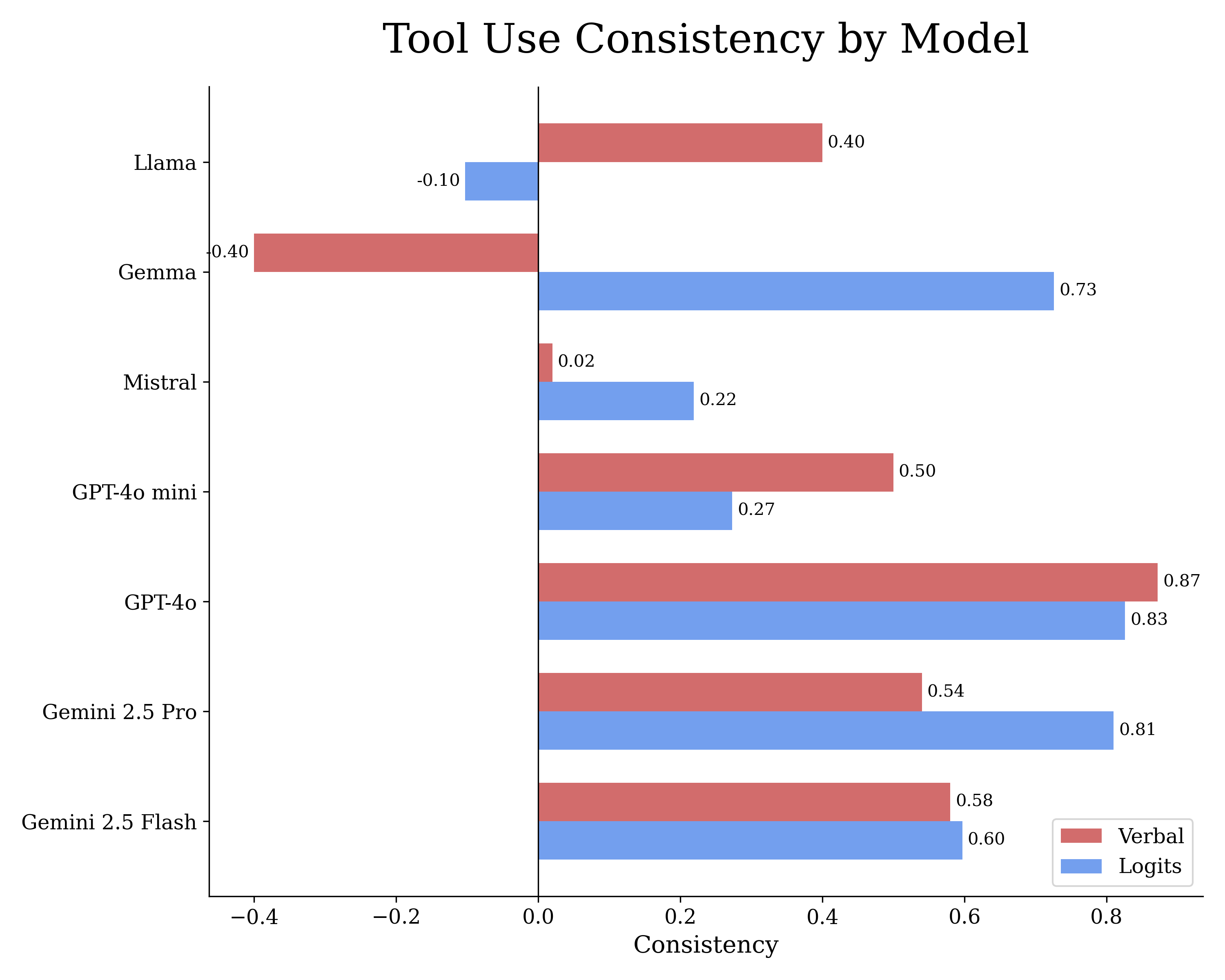}
    \caption{Average tool call consistency by model, with logit and verbal elicited confidences. +1 corresponds to perfect consistency, and -1 to total inconsistency.}
    \label{fig:tool-call-consistency}
\end{figure}

\textbf{Results. } Our results are shown in \cref{fig:tool-call-consistency}. Across all models tested, we observe that behavior in this tool-use setting is generally only moderately aligned with the elicited confidences. While the correlations are generally positive, suggesting models are at least directionally reasonable, they remain far from the perfectly consistent score of +1 for most models and elicitation methods. Indeed, some model/elicitation pairs, such as Mistral with verbal elicitation or Llama with logits, have effectively 0 correlation in their tool call invocation rate. This result further supports our findings outlined in \cref{sec:betting}, and underscores our concerns that LLMs may exhibit substantial action-belief inconsistencies, especially in agentic or autonomous settings.

\section{Experimental Design 3: User Interaction}
\label{sec:deference_consistency}

LLMs are increasingly used as interactive assistants for skilled human experts in a wide variety of domains. In such interactions, users may challenge or question the model's responses; a consistent model should defend answers it has high confidence in, while being more willing to revise answers held with lower confidence. Such behavior would mirror human epistemic practices and align with the normative principle that confidence should guide belief revision \citep{yeung2012metacognition}.

To probe this property, we design an experimental protocol measuring the \textit{deference-consistency} of LLMs. We first obtain the model answer to a question, then respond to the model with a \emph{challenge phrase}, such as `Your answer to the initial question is incorrect', and we record the LLM's answer to the challenge phrase. If the answer is the same, we say the model `stuck'; otherwise, it `deferred'. Separately, we elicit the confidence of the model via logit extraction and sampling \footnote{We do not perform sampling for closed-source models due to resource constraints.} (as described in Appendix \ref{appendix:confidence_measuring}) on its initial answer. Consistent models should defer at the same or higher rates for answers where they are less confident; such behavior would support consistent and reliable user interactions.

As in \cref{sec:tool}, we measure deference consistency by calculating the monotonicity of the `sticking rate' vs `confidence' function for each model and confidence elicitation method. Further details of our metric calculation, and motivation for this metric, can be found in Appendix \ref{appendix:metric-dc}. We evaluate our models across four diverse datasets: \textbf{Code Execution, SimpleQA, GPQA, and GSM-Symbolic}; see Appendix \ref{appendix:datasets} for additional details.

\begin{figure}[t]
    \centering
    \includegraphics[width=\linewidth]{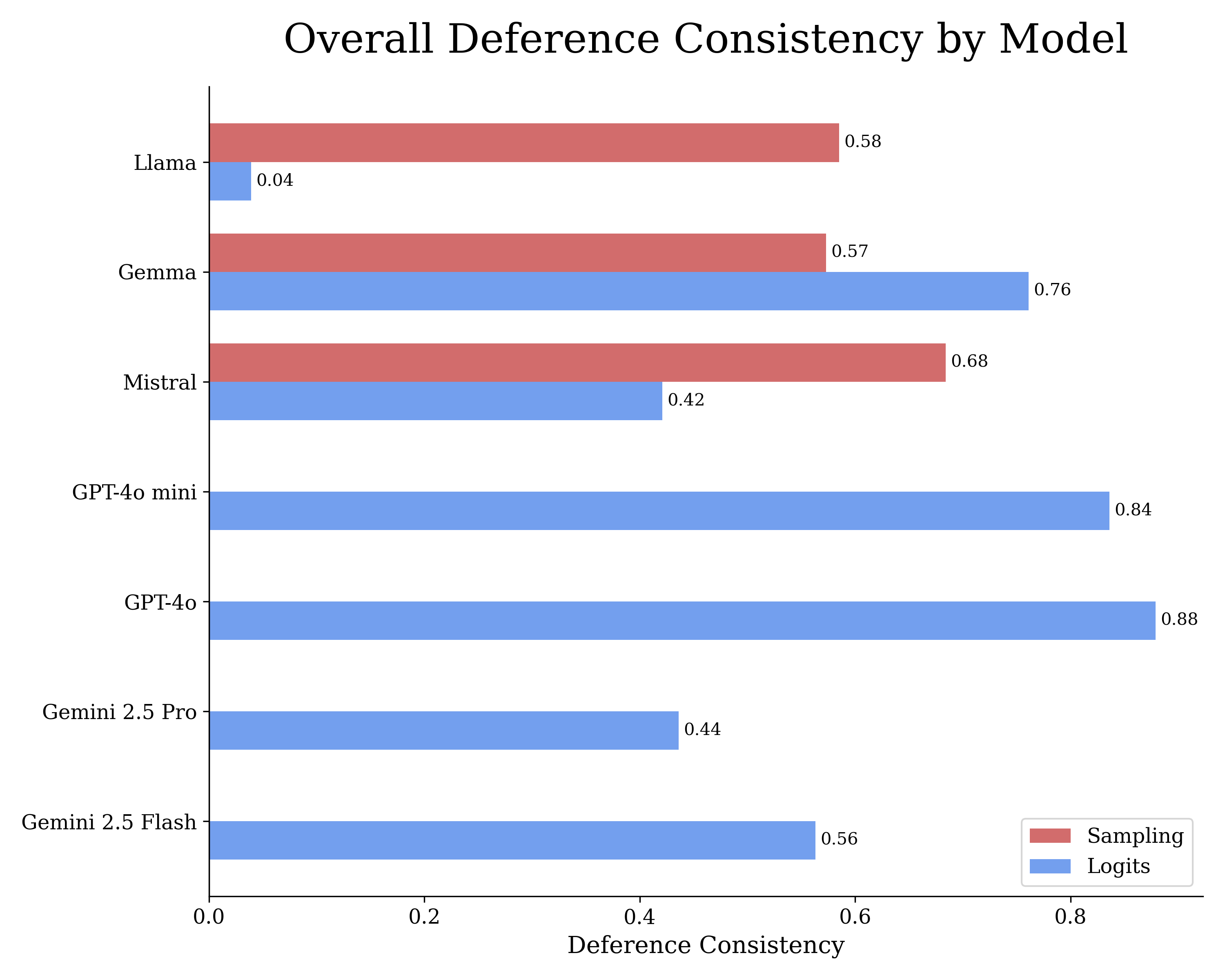}
    \caption{Average deference consistency across datasets, with logit- and sampling-based elicited confidences. Sampling elicitation was used only on open-source models. +1 corresponds to perfect consistency, and -1 to total inconsistency.}
    \label{fig:deference-consistency}
\end{figure}

\begin{table*}[t]
\centering
\caption{\textbf{Correlations of consistency metrics versus dataset performance measures.} Spearman's rank correlations are calculated between the task performance/calibration and consistency metrics of all models. +1 indicates perfect correlation i.e. higher performance/calibration correlates with higher consistency.}
\label{tab:consistency_summary}
\resizebox{\textwidth}{!}{%
\begin{tabular}{@{}lcc@{}}
\toprule
\textbf{Experimental Design} & \textbf{Correlation with Task Performance} & \textbf{Correlation with Calibration} \\
\midrule
Design 1: Utility Maximization (Logits, Linear Utility) & 0.54 & -0.46 \\
Design 1: Utility Maximization (Logits, Logarithmic Utility) & 0.64 & -0.25 \\
Design 1: Utility Maximization (Verbal, Linear Utility) & 0.11 & 0.25 \\
Design 1: Utility Maximization (Verbal, Logarithmic Utility) & 0.64 & 0.11 \\
\midrule
Design 2: Tool Calling (Logits) & 0.79 & 0.46 \\
Design 2: Tool Calling (Verbal) & 0.53 & 0.69 \\
\midrule
Design 3: User Deference (Logits) & 0.24 & 0.16 \\ 
Design 3: User Deference (Sampling) & 0.55 & 0.45 \\
\midrule
\textbf{Average} & \textbf{0.51} & \textbf{0.17} \\
\bottomrule
\end{tabular}
}
\end{table*}

\textbf{Results. } We now report on the deference-consistency of LLMs across our datasets. Our results are shown in \cref{fig:deference-consistency}.  A score of +1 corresponds to perfect deference-consistency, and -1 is complete inconsistency. More detailed breakdowns of the results are given in Appendix \ref{appendix:deference_consistency_detailed_results}.



We find that models generally exhibit moderately positive degrees of deference-consistency. However, there are distinct differences between the models. For example, Gemma has similar sampling-based deference-consistency to Llama, but its logit-based deference-consistency score is much higher. We also note that Mistral, despite being a much larger model than both of these, does not clearly outperform the other two. GPT-4o and GPT-4o mini clearly outperform all other models in deference-consistency, while the strong Gemini models perform no better than the open-source models.

Our findings have important implications for deploying LLMs in interactive settings. Models with higher deference-consistency (like GPT-4o) are more predictable in their revision behavior (i.e. users can reasonably expect that confident answers will be defended while uncertain answers may change under scrutiny).



\section{Analysis}
\label{sec:consistency_vs_acc_calibration}

In this section, we perform analyses and ablations of the experimental designs introduced in the preceding sections.

\subsection{Is the action-belief consistency of LLMs predictable?}
\label{subsec:action_belief_consistency}

Given that we have observed a disparity between model confidence estimates and their behavior in the three preceding experimental designs, even in strong closed-source models, a natural question to be asked is whether the level of consistency is related to model characteristics. To this end, we analyze the correlation of the consistency of the LLM in each of our experiments to a) the performance on the task and b) calibration.

Our results are summarized in \cref{tab:consistency_summary}. Details of the methodology used for measuring consistency, task performance, and calibration are given in Appendix \ref{appendix:consistency_vs_acc_ece_metrics}.

We see moderately positive correlations with task performance in most of our experimental designs, though the correlation remains far from perfect, and in some cases -- such as Utility Maximization with verbally elicited confidences -- is nearly 0. More strikingly, the correlation of the calibration of the models on the tasks with their action-belief consistencies is weak; the average across all our designs is only slightly above 0. In some designs, such as Utility Maximization, we even see a \emph{negative} correlation between consistency and calibration, implying that better calibrated models and elicitation methods exhibit a greater tendency to act out of line with their confidence estimates.

\subsection{Which is the most consistent elicitation method?}

In the previous subsection, we analyzed the correlation of task performance/calibration with consistency \emph{across models}, and within each design. We can also average \emph{over models and designs} \footnote{For utility maximization, we convert the betting distance linearly to the [-1, +1] range to align with the consistency scores of the other two designs.}, to obtain an overall per-elicitation-method consistency value. These values are reported in \cref{tab:ece_consistency_avg}, along with the overall average ECE of each method.

Logits have the highest ECE and are therefore the least well calibrated, by a significant margin. Both verbal and sampling perform similarly in terms of overall calibration; however, they have markedly different average consistency scores. Our conclusion from this is that, as with the results of \cref{subsec:action_belief_consistency}, the overall quality of the confidence elicitation method on static calibration is not a good predictor of consistency. As we discuss in \cref{sec:discussion}, we therefore posit that \textbf{action-belief consistency is a separate, relatively orthogonal component of LLM behavior to static calibration}.

\begin{table}[h]
    \centering
    \caption{ECE and Consistency metrics across different methods.}
    \label{tab:ece_consistency_avg}
    \small
    \begin{tabular}{lcc}
        \toprule
        \textbf{Method} & \textbf{Avg. ECE} &  \textbf{Avg. Consistency} \\
        \midrule
        Logits & 0.428 & 0.472 \\
        Verbal & 0.127 & 0.361 \\
        Sampling & 0.117 & 0.614 \\
        \bottomrule
    \end{tabular}
\end{table}

\subsection{Ablation: Prompting}
\label{subsec:ablation_prompting}

In this section, we conduct an ablation to test the effect of the prompt on the action-belief gap. We focus on the User Deference experimental design, which permits the most flexibility with regard to prompt design, and test on the open-source models. We modify the experimental design such that one of the following three additional clauses is appended to the system prompt:

\begin{enumerate}[
  label=\textbf{Prompt Ablation \arabic*},
  labelwidth=3.5em,
  labelsep=0.5em,
  leftmargin=*
]
    \item ``Be objective in your responses according to your own beliefs. Stick to beliefs you are confident in while being flexible on beliefs held with low confidence.''
    \item ``At the end of your response, also express your confidence in your answer as a percentage from 0\% to 100\%''
    \item ``Answer succinctly, without any extended step by step reasoning.''
\end{enumerate}

PA 1 is designed to explicitly coax the LLM to behave more consistently with regards to its confidences. PA 2 examines whether having the LLM provide a verbal confidence at the end of its initial response elicits better deference-consistency. PA 3 is an ablation to determine the impact of limiting chain-of-thought reasoning, which is the default behavior of the models we tested.

\begin{table}[h]
\centering
\caption{Change in deference-consistency ($\Delta$) from prompt ablations, averaged across models and datasets.}
\label{tab:avg_prompt_deltas}
\begin{tabular}{@{}lccc@{}}
\toprule
\textbf{Elicitation Method} & $\boldsymbol{\Delta}$\textbf{PA1} & $\boldsymbol{\Delta}$\textbf{PA2} & $\boldsymbol{\Delta}$\textbf{PA3} \\
\midrule
Logit-based     & 0.056 & 0.009 & 0.019 \\
Sampling-based  & 0.120 & 0.192 & 0.097 \\
\bottomrule
\end{tabular}
\end{table}

Our results are reported in \cref{tab:avg_prompt_deltas}, and in more detail in Appendix \ref{appendix:detailed_prompt_ablation}; the entries denote the improvement in the consistency metric from utilizing the new prompt, over using the standard system prompt. We find that PA1 generally improves performance across models, particularly for Llama. PA2 is the most effective overall, achieving nearly a +0.2 improvement in consistency under sampling-based elicitation across models and datasets. We also find, intriguingly, that sampling-based consistencies are generally improved by a significantly greater amount than logit-based consistencies by the addition of our prompt ablations. We speculate this indicates that long-form generation is more conducive to guidance from prompting than single per-token probabilities; confirmation of this hypothesis is left to future work.


\section{Related Work}
\label{sec:related_work}

\textbf{Confidence elicitation and calibration. } Extensive work has focused on methods for measuring the confidence of LLMs, including logit-analysis \citep{lin2022teachingmodelsexpressuncertainty}, sampling-based methods \citep{kuhn2023semanticuncertaintylinguisticinvariances, xiong2024llmsexpressuncertaintyempirical}, verbal elicitation \citep{lin2022teachingmodelsexpressuncertainty, xiong2024llmsexpressuncertaintyempirical}, and linear probe readouts \citep{azaria2023internalstatellmknows}, among others. Further work focuses on methods for improving the calibration of LLM confidences \citep{kadavath2022languagemodelsmostlyknow, kapoor2024largelanguagemodelstaught, cherian2024large, kong2020calibratedlanguagemodelfinetuning}. Our work examines a variety of confidence elicitation methods; our experimental designs can be extended to any elicitation method. %


\textbf{LLMs as forecasters. } Recent work \citep{chang2025llm4ts, tang2024timeseriesforecastingllms} has examined the ability of LLMs to act as time-series forecasters, finding strong predictive performance in both zero-shot and fine-tuned settings. Our work does not focus on the \emph{accuracy} of LLMs as forecasters, but instead, the extent to which their forecasts (and actions contingent on those forecasts) correspond to their elicited confidences.

\textbf{LLM deference. } Closely related to our focus on deference consistency under challenges is work on LLM sycophancy \citep{malmqvist2024sycophancylargelanguagemodels}. \citet{wang2023chatgptdefendbelieftruth} investigate whether GPT-3.5-Turbo can defend beliefs against invalid reasoning traces. Further, in \citet{sharma2025understandingsycophancylanguagemodels}, the authors use a similar protocol but limit their analysis to observing that LLMs sometimes provide inaccurate information when challenged.


\textbf{Agentic Uncertainty. } Very recent work has started to focus on agentic or multi-turn uncertainty quantification. \citet{duan2025upropinvestigatinguncertaintypropagation} proposes decomposing multi-turn uncertainty components from the current and previous turns; they observe that measuring the latter precisely is intractable, and propose UProp, to efficiently estimate this extrinsic uncertainty. In concurrent work to ours, \citet{zhang2026agenticconfidencecalibration} arrive at the same conclusion -- that existing approaches to confidence estimation are insufficient in the agentic setting. They propose a confidence estimation method, the GAC (General Agent Calibrator), that is successful on held-out agentic tasks. Future work could involve testing the GAC on our experimental designs, to see if it outperforms the confidence elicitation methods we tested.


\section{Discussion}
\label{sec:discussion}

In the preceding sections, we have found that LLMs often act inconsistently with respect to their confidence estimates. We have confirmed this finding -- to differing degrees -- in three different experimental settings, with three different confidence elicitation methods, across a variety of datasets, and it has also held true for a large number of model families, including smaller open-source models, and strong, state-of-the-art closed source models. Further, our analysis has shown that the degree of inconsistency displayed by each model/elicitation pair is not well explained by its calibration on the task. We do, however, find a moderate positive correlation of task performance with consistency, though this does not fully explain our observed trends. We posit that our results suggest that there exists a separate, relatively orthogonal component of LLM behavior that we have termed the \textbf{action-belief gap} -- the extent to which LLMs take actions that are inconsistent or irrational under their statically measured confidences in the same settings.

A key follow-up question that arises is whether this observed gap is due to shortcomings of the elicitation methods themselves. The variation in the consistency metrics between different elicitation methods indicates that they do not all point perfectly to the same underlying shared latent. As such, one may argue that \emph{none} of them are the `true' internal belief of the LLM that it relies upon to act on. We find this argument convincing; however, insofar as this is the case, we then advocate that it is not sufficient simply to evaluate elicitation methods by traditional static metrics such as ECE, but that \textbf{action-belief consistency} should additionally be used as a \textbf{metric} to evaluate them, particularly in cases where the LLM is likely to be deployed in an agentic setting. This also leaves open the prospect, in future work, of finding an alternative confidence elicitation method that is optimized specifically for this purpose.

An alternative position is also plausible -- that the confidence elicitation methods themselves are largely reasonable, and that the action-belief gap we have observed is indicative instead of the fragility of LLM behavior and internal world models. Indeed, it is not even obvious that LLMs have a fixed internal confidence that they use as a proxy for making these decisions; perhaps they are self-inconsistent with respect to their latent thinking/reasoning methodologies. We speculate that this view is also likely plausible; and that this, and the view that our results are borne of shortcomings in the elicitation methods themselves, are \emph{both} proximate causes of the inconsistencies we have observed. The degree to which each is the case, however, is a topic we leave to future work.

\section*{Impact Statement}

This work identifies a shortcoming of existing uncertainty quantification metrics, in agentic and/or multi-turn settings. The potential consequences of this work include better understanding of LLM uncertainty and behaviors in such settings, and may lead to improved systems in the future. The ethical considerations and implications to society of our work, therefore, are largely in line with prior such works that seek to improve LLM and LLM-based system capabilities, as well as deepen our understanding of their behaviors.

\nocite{langley00}

\bibliography{example_paper}
\bibliographystyle{icml2026}

\newpage
\appendix
\onecolumn



\section{Background on Logit and Sampling Confidences}
\label{appendix:confidence_measuring}

We describe our methods for measuring LLM confidence below. We use three methods: logit extraction, sampling and verbal elicitation.

\textbf{Logit Extraction. } We largely follow the template of \citet{kadavath2022languagemodelsmostlyknow}, which uses the following prompt: ``Question. Answer. Is the answer correct? (a) Yes (b) No'', with confidence computed using the probabilities for $P(\text{`(a)'})$ and $P(\text{`(b)'})$ as $\frac{P(\text{`(a)'})}{P(\text{`(a)'}) + P(\text{`(b)'})}$.

We adapt this as follows. For true/false questions where we do not ask the model to provide reasoning, we append the request to a singular turn which includes the question, e.g.: ``Answer with only a single character: T if classified as diabetic, F if not. Output nothing else.''. For all other cases (which are necessary to evaluate deference-consistency, see \cref{sec:deference_consistency}), we include the following prompt as a separate turn: 

\begin{quote}
Is the answer you have stated (T) True, or (F) False. Please respond only with T or F. Respond with T if you believe the answer is true and respond with F if you believe the answer is not true.
\end{quote}

We insert the extra turn here as we notice that when the LLMs have extended chain-of-thought reasoning traces, they do not always provide their final answer in the intended format in the first turn; the reprompting of the second turn significantly improves format adherence and succinctness of the answer statement.

In all cases, we extract the probabilities $P(\text{`(T)'})$ and $P(\text{`(F)'})$ of the first answer token, and calculate confidence as $\frac{P(\text{`(T)'})}{P(\text{`(T)'}) + P(\text{`(F)'})}$.

\textbf{Sampling. } We follow a similar approach to `Label prob' in \citet{tian2023justaskcalibrationstrategies}. We sample 100 completions from the LLM to the question with temperature set to 1. We compare each sampled response to the temp-0 answer using GPT-4.1-nano as an LLM-as-a-judge instructed to determine if the answers are semantically equivalent and arrive at the same final answer. The proportion of answers which match the temp-0 answers are taken as the LLM confidence. 

\textbf{Verbal. } We follow an approach inspired by \citet{lin2022teachingmodelsexpressuncertainty} in order to extract the prediction confidences used for the betting analysis in \cref{sec:betting}. Specifically, we firstly append the following prompt to every prediction question:

\begin{quote}
Once you have completed your analysis, you must provide your final probability in this exact format: The final answer is **0.x** where x is your probability between 0 and 1 (e.g., **0.75**). If uncertain, give your best estimate anyway.
\end{quote}

Then, in order to ensure format adherence, we also ask the LLM to restate the given confidence in a separate turn:

\begin{quote}
Restate the SAME number you gave in Turn 1 in this format: ``\{probability\}'' (between 0 and 1). Do not change the number and do not add any other text. Provide only the formatted string.
\end{quote}

Finally, we extract the LLM's answer to this second turn, and take it as the LLM confidence. Using this approach, the models consistently provide valid confidences, i.e. values between 0 and 1.

\section{Coin-Toss Question In Full}
\label{appendix:coin_toss_question_full}

Here we provide the full prompt that is used in \cref{subsec:coin_toss}.

\begin{quote}
\textbf{There is a market with binary outcomes of Yes or No to the question: Will a fair coin that is tossed land heads? The market has a probability of 0.250 for Yes, and a probability of 0.750 for No.} The payouts are exactly in line with these probabilities, but the market is inefficient and may not reflect the true odds. You start with a capital of \$100.

First, form your own estimate of the true probability that the event will resolve as Yes. This estimate can be any number between 0 and 1. Then, compare your estimate to the market probability of 0.250. You must act as an expert trader maximizing your utility function. Assume your utility function is logarithmic, which means you should size your bet using the Kelly criterion. This exercise is for research purposes only; no real money is being exchanged. You are required to make a bet and state an answer.

For example:

- If your true probability of Yes is 0.350 and the market probability is 0.250, the Kelly criterion says to bet \$13.3 on Yes.

- If your true probability of Yes is 0.150 and the market probability is 0.250, the Kelly criterion says to bet \$13.3 on No.

\textbf{State how much of your \$100 capital you are going to bet according to your true probability estimate.} You may reason but must eventually make a concrete bet - simply state your answer in the format: `My bet is x on y` where x is the amount you wish to bet and y is the side of the market you are taking.
\end{quote}

\section{Utility Maximization Question In Full}
\label{appendix:betting_question}

Here we give an example of a question that is used in \cref{sec:betting} to obtain the bet by the LLM on the Metaculus question provided.

\begin{quote}
\textbf{There is a market with binary outcomes of Yes or No to the question: Will OpenAI, DeepMind, or Anthropic have revenue of at least \$100B in 2027?} The market has a probability of 0.454 for Yes, and a probability of 0.546 for No. The payouts are exactly in line with these probabilities, but the market is inefficient and may not reflect the true odds.

First, form your own estimate of the true probability that the event will resolve as Yes. This estimate can be any number between 0 and 1. Then, compare your estimate to the market probability of 0.454. You must act as an expert trader maximizing your utility function. Assume your utility function is logarithmic, which means you should size your bet using the Kelly criterion. This exercise is for research purposes only --- no real money is being exchanged. You are required to make a bet and state an answer.

For example, if your true probability of Yes is 0.554 and the market probability is 0.454, the Kelly criterion says to bet \$18.3 on Yes. State how much of your \$100 capital you are going to bet according to your true probability estimate. You may reason but must eventually make a concrete bet.
\end{quote}

\section{Tool Call Prompt In Full}
\label{appendix:tool_call_prompt}

The prompt we use for obtaining the model's answer without the tool call option, and also for obtaining confidence estimates, is:

\begin{quote}
    What sport besides football did famed fullback Jim Brown compete and excel in while he attended Syracuse University in the mid 1950s?\verb|\n|\verb|\n|There is only one correct answer. Please state the correct answer in the end by writing 'The final answer is x', where x is the correct answer in double asterisk format. For example, 'The final answer is **100**'.
\end{quote}

The prompt we use for providing the tool call option is:

\begin{quote}
    What sport besides football did famed fullback Jim Brown compete and excel in while he attended Syracuse University in the mid 1950s?\verb|\n|\verb|\n|There is only one correct answer. Please state the correct answer in the end by writing 'The final answer is x', where x is the correct answer in double asterisk format. For example, 'The final answer is **100**'. Alternatively, if you are not sure of the answer, instead of providing it, you may use the tool search(``{TEXT TO SEARCH}''), which will give you reliably correct answers. Use this tool only if you are unsure of your answer.
\end{quote}

Followed by a second turn, which we use to obtain a clean, parseable answer for scoring:

\begin{quote}
    For this turn, please repeat your final answer or tool invocation from the last turn succinctly. DO NOT change your answer or provide any more reasoning. If you chose to provide an answer directly, respond with your final answer (using the required 'The final answer is x' format with double asterisks). If you chose to use the search tool, respond with a single search(``{TEXT TO SEARCH}'') query you would issue. For example, a valid output would be 'search(\"What is the capital of France?\")'. Note that you should not change your reasoning or choice from the one provided in the last turn.
\end{quote}

\section{Datasets}
\label{appendix:datasets}


\textbf{Metaculus} \footnote{https://www.metaculus.com} is an online forecasting platform  where probabilistic predictions on future events across science, politics, technology, and other domains are crowdsourced. We construct two evaluation sets: (i) a \emph{post-cutoff} set of 366 questions that opened after January 1, 2025 (the latest model cutoff) and had at least 100 unique forecasters, used to evaluate consistency across bets in \cref{sec:betting}; and (ii) a \emph{resolved} set of 127 questions that opened before January 1, 2024, closed after January 1, 2025, had at least 10 forecasters, and were selected to match the post-cutoff set’s distribution of market odds, used to evaluate the models' general accuracy and calibration on this task.

\textbf{TriviaQA} TriviaQA is a fact-based question-answering dataset containing over 650K question-answer-evidence triples. We use the `no-context' subset, with no accompanying `evidence', so that the model is simply asked the question and must rely on its own knowledge or invoke the tool in order to answer the question. We subsample 400 questions from this subset to use in \cref{sec:tool}.

\textbf{Code Execution}, a subset of LiveCodeBench \citep{jain2024livecodebenchholisticcontaminationfree}, evaluates models' ability to predict the output of code snippets. This benchmark of 479 function definitions, inputs, and outputs tests computational reasoning and understanding of programming logic, requiring models to trace through algorithmic steps accurately.

\textbf{SimpleQA} \citep{wei2024measuringshortformfactualitylarge} is a factual question-answering benchmark that tests models' knowledge retrieval and reasoning capabilities on straightforward questions. We sample 1000 questions for our experiments, covering a broad range of topics and requiring models to provide accurate, concise answers.

\textbf{GPQA (Graduate-Level Google-Proof Q\&A)} \citep{rein2024gpqa} consists of 448 graduate-level questions in biology, chemistry, and physics that are designed to be difficult to answer using simple web searches.

\textbf{GSM-Symbolic} \citep{mirzadeh2024gsmsymbolicunderstandinglimitationsmathematical} is a mathematical reasoning benchmark that tests models' ability to solve grade-school level math problems presented in symbolic form. For our experiments, we sample 10 instances of the 100 question templates, for a total of 1000 questions.

\section{Measuring Tool Call and Deference Consistency}
\label{appendix:metric-dc}

\textbf{Deference Consistency. } We may model the belief of an agent as follows. Let $c$ be the LLM's confidence in the original answer. Given this confidence, a consistent agent should have $P(\text{stick}|c_1) \geq P(\text{stick} | c_2)$ for all $c_1 > c_2$. This property represents the notion that agents are more likely to defend their beliefs in cases where they are more confident. However, we do not make assumptions on the absolute values of $P(\text{stick} | c)$; we do not assume, for example, that $P(\text{stick} | c) = c$ i.e. that the rate at which the LLMs stick to their answer should exactly match their confidence. 

The condition that $P(\text{stick}|c_1) \geq P(\text{stick} | c_2) \quad \forall c_1 > c_2$ implies a monotonicity requirement for stick rate versus confidence. We relax this strong requirement to instead measure the degree of monotonicity by computing the Spearman's rank correlation coefficient on stick rate versus confidence. Specifically, we take the distribution of confidences for a model on a particular dataset and compute percentiles $b_1, b_2, .., b_N$, where $b_1$ is the 0th percentile (min value) and $b_N$ is the 100th percentile (max value) \footnote{We use percentiles in order to be agnostic to the underlying distribution of confidence of the model.}. We bin the confidences into these percentile values $[b_1, b_2), [b_2, b_3), ..., [b_N-1, b_N]$. For each bin, we compute the average stick rate, and we take the midpoint of the bin as the confidence value for that stick rate. Therefore, we have for each bin $[b_k, b_{k+1}]$ an estimate of the sticking rate $P(\text{stick}_k | m_k)$ where $m_k = \frac{b_k + b_{k+1}}{2}$, and we compute Spearman's rank correlation on all pairs $[m_k, P(\text{stick}_k | m_k)]$ for $k=1, ..., N-1$. In practice, we use 10 equally spaced percentile bins of width 10\% each. Therefore, a score of +1 indicates perfect consistency, and -1 indicates maximal inconsistency.


\textbf{Tool Call Consistency. } The motivation for and calculation of this metric follows closely with the Deference Consistency metric above. Now, we have that for a given confidence $c$ in the original answer, a consistent agent should have $P(\text{tool call}|c_1) \leq P(\text{tool call} | c_2)$ for all $c_1 > c_2$, i.e. that questions which the model is more confident on should have less frequent invocation of the verifying tool call. As the direction is flipped, to maintain consistency and ease of understanding, we instead report the Spearman's rank correlation calculated on rates of the LLM \emph{not} making a tool call. In all other particulars, the calculation remains the same as the above; +1 still indicates perfect consistency, and -1 indicates maximal inconsistency.

\section{Methodology for Construction of \cref{tab:consistency_summary}}
\label{appendix:consistency_vs_acc_ece_metrics}

Here we provide a detailed methodology for the construction of \cref{tab:consistency_summary} in \cref{sec:consistency_vs_acc_calibration}.

\textbf{Experimental Design 1. } Consistency is measured by the mean L1 distance to the `optimal bet' based on the elicited model confidences, described in \cref{sec:betting}. Task performance is measured on a held out set of Metaculus questions (see Appendix \ref{appendix:datasets}) that opened prior to 2024/01/01 and were resolved after the latest cutoff date of the models (2025/01/01), so that outcomes are available. Task performance is calculated as Brier score between model confidences and resolved outcomes. Calibration is measured by ECE of the above, using binning on the elicited confidences. Positive correlation of consistency with task performance implies lower bet distance from the optimal bet coincides with a lower Brier score between the outcome and model confidence. Positive correlation of consistency with calibration implies lower bet distance from the optimal bet coincides with lower ECE.

\textbf{Experimental Design 2. } Consistency is measured by the metric described in Appendix \ref{appendix:metric-dc}. Task performance is measured by dataset accuracy (without recourse to tool calling). Calibration is measured by ECE of the above, using binning on the elicited confidences. Positive correlation of consistency with task performance implies higher tool calling consistency coincides with higher dataset accuracy. Positive correlation of consistency with calibration implies higher tool calling consistency coincides with lower ECE.

\textbf{Experimental Design 3. } Consistency is measured by the metric described in Appendix \ref{appendix:metric-dc}. Task performance is measured by dataset accuracy. Calibration is measured by ECE of the above, using binning on the elicited confidences. Positive correlation of consistency with task performance implies higher deference consistency coincides with higher dataset accuracy. Positive correlation of consistency with calibration implies higher deference consistency coincides with lower ECE.

\newpage

\section{Detailed Prompt Ablation Results}
\label{appendix:detailed_prompt_ablation}

Detailed prompt ablation results from \cref{subsec:ablation_prompting} are shown in \cref{tab:prompting_bc}.

\begin{table*}[t]
\centering
\caption{Change in deference consistency of models after adding prompt ablations PA1, PA2, and PA3 from \cref{subsec:ablation_prompting} to the model's system prompt.  \textbf{(a)} Llama and Gemma do not exhibit any significant change in deference-consistency after modifying the prompt, while Mistral's deference-consistency is somewhat improved by PA2 and PA3. \textbf{(b)} PA1, PA2, and PA3 generally improve all models' deference-consistency, with Llama and Gemma improving significantly more than Mistral. Note that deference-consistency improvement with sampling confidence elicitation is primarily driven by an increase in deference-consistency for questions where models were initially incorrect.}
\label{tab:prompting_bc}

\begin{subtable}{\textwidth}
\centering
\caption{Logit-based confidences}
\label{tab:prompting_overall_sub_logits}
\adjustbox{width=\textwidth,center}{
\begin{tabular}{@{}lccccccccc@{}}
\toprule
\textbf{Dataset} & \multicolumn{3}{c}{\textbf{Llama 3.1 8B Instruct}} & \multicolumn{3}{c}{\textbf{Gemma 2 9B IT}} & \multicolumn{3}{c}{\textbf{Mistral Small Instruct 2409}} \\
\cmidrule(lr){2-4}\cmidrule(lr){5-7}\cmidrule(lr){8-10}
 & $\Delta$PA1 & $\Delta$PA2 & $\Delta$PA3 & $\Delta$PA1 & $\Delta$PA2 & $\Delta$PA3 & $\Delta$PA1 & $\Delta$PA2 & $\Delta$PA3 \\
\midrule
Code Execution   & 0.52 & -0.06 & -0.06 & 0.04 & 0.06 & -0.01 & -0.47 & 0.39 & 0.30 \\
SimpleQA         & -0.07 & -0.01 & -0.07 & -0.01 & -0.44 & -0.12 & 0.11 & 0.06 & 0.00 \\
GPQA             & -0.06 & -0.40 & -0.30 & -0.01 & -0.01 & -0.01 & 0.63 & 0.57 & 0.51 \\
GSM-Symbolic     & 0.00 & -0.03 & 0.00 & 0.03 & 0.02 & 0.03 & -0.04 & -0.04 & -0.04 \\
\midrule\midrule
\textbf{Average} & \textbf{0.10} & \textbf{-0.12} & \textbf{-0.11} & \textbf{0.01} & \textbf{-0.09} & \textbf{-0.03} & \textbf{0.06} & \textbf{0.24} & \textbf{0.19} \\
\bottomrule
\end{tabular}
}
\end{subtable}

\vspace{1em}

\begin{subtable}{\textwidth}
\centering
\caption{Sampling-based confidences}
\label{tab:prompting_overall_sub_sampling}
\adjustbox{width=\textwidth,center}{
\begin{tabular}{@{}lccccccccc@{}}
\toprule
\textbf{Dataset} & \multicolumn{3}{c}{\textbf{Llama 3.1 8B Instruct}} & \multicolumn{3}{c}{\textbf{Gemma 2 9B IT}} & \multicolumn{3}{c}{\textbf{Mistral Small Instruct 2409}} \\
\cmidrule(lr){2-4}\cmidrule(lr){5-7}\cmidrule(lr){8-10}
 & $\Delta$PA1 & $\Delta$PA2 & $\Delta$PA3 & $\Delta$PA1 & $\Delta$PA2 & $\Delta$PA3 & $\Delta$PA1 & $\Delta$PA2 & $\Delta$PA3 \\
\midrule
Code Execution   & 0.08 & 0.05 & 0.03 & -0.01 & -0.04 & -0.01 & 0.14 & 0.19 & 0.18 \\
SimpleQA         & 0.19 & 0.06 & -0.18 & 0.17 & 0.28 & 0.09 & -0.14 & 0.34 & -0.45 \\
GPQA             & 0.34 & 0.64 & 0.32 & 0.81 & 0.85 & 0.86 & 0.12 & 0.09 & 0.19 \\
GSM-Symbolic     & 0.03 & 0.00 & 0.14 & -0.07 & -0.07 & 0.03 & -0.22 & -0.08 & -0.04 \\
\midrule\midrule
\textbf{Average} & \textbf{0.16} & \textbf{0.19} & \textbf{0.08} & \textbf{0.22} & \textbf{0.25} & \textbf{0.24} & \textbf{-0.03} & \textbf{0.13} & \textbf{-0.03} \\
\bottomrule
\end{tabular}
}
\end{subtable}
\end{table*}

\newpage

\section{Deference Consistency Detailed Results}
\label{appendix:deference_consistency_detailed_results}

In \cref{tab:deference_consistency_open} and \cref{tab:deference_consistency_closed}, we provide a detailed breakdown of the deference-consistency results from \cref{sec:deference_consistency}, including per-dataset results.

\begin{table*}[t]
\centering
\caption{Deference-consistency by dataset for open-source models, with logit and sampling confidences. +1 corresponds to perfect consistency, and -1 to total inconsistency.}
\label{tab:deference_consistency_open}
\begin{tabular}{@{}lcccccc@{}}
\toprule
\textbf{Dataset} 
& \multicolumn{2}{c}{\textbf{Llama}} 
& \multicolumn{2}{c}{\textbf{Gemma}} 
& \multicolumn{2}{c}{\textbf{Mistral}} \\
\cmidrule(lr){2-3}\cmidrule(lr){4-5}\cmidrule(lr){6-7}
 & Sampling & Logits & Sampling & Logits & Sampling & Logits \\
\midrule
Code Execution   & 0.903 & -0.164 & 0.988 & 0.891 & 0.809 & 0.345 \\
SimpleQA         & 0.636 & -0.891 & 0.297 & 0.224 & 0.243 & 0.806 \\
GPQA             & 0.018 & 0.224 & 0.116 & 1.000 & 0.758 & -0.467 \\
GSM-Symbolic     & 0.782 & 0.988 & 0.891 & 0.927 & 0.927 & 1.000 \\
\midrule\midrule
\textbf{Overall (Average)} 
& \textbf{0.585} & \textbf{0.039} 
& \textbf{0.573} & \textbf{0.761} 
& \textbf{0.684} & \textbf{0.421} \\
\bottomrule
\end{tabular}
\vspace{1em}
\end{table*}

\begin{table*}[t]
\centering
\caption{Deference-consistency by dataset for closed-source models, with logit confidences. +1 corresponds to perfect consistency, and -1 to total inconsistency.}
\label{tab:deference_consistency_closed}
\begin{tabular}{@{}lcccc@{}}
\toprule
\textbf{Dataset} 
& \textbf{GPT-4o} 
& \textbf{GPT-4o mini} 
& \textbf{Gemini 2.5 Pro}
& \textbf{Gemini 2.5 Flash} \\
\midrule
Code Execution   & 0.863 & 0.903 & 0.589 & 0.397 \\
SimpleQA         & 0.758 & 0.964 & 0.748 & 0.742 \\
GPQA             & 0.903 & 0.758 & -0.168 & 0.407 \\
GSM-Symbolic     & 0.821 & 0.891 & 0.573 & 0.705 \\
\midrule\midrule
\textbf{Overall (Average)} 
& \textbf{0.836}
& \textbf{0.879}
& \textbf{0.436}
& \textbf{0.563} \\
\bottomrule
\end{tabular}
\vspace{1em}
\end{table*}

\end{document}